\documentclass[conference]{IEEEtran}
\ifCLASSINFOpdf
\else
\fi

\usepackage{float}
\usepackage{graphicx}
\usepackage{booktabs}
\usepackage{array}
\usepackage{amsmath}
\usepackage{tabu}
\usepackage[export]{adjustbox}
\usepackage{subcaption}
\usepackage{amssymb}
\usepackage{authblk}

\begin{document}

\title{Aiding Intra-Text Representations with \\Visual Context for Multimodal \\Named Entity Recognition}


\author[*]{Omer Arshad}
\author[$\dag$]{Ignazio Gallo}
\author[ $\dag$]{Shah Nawaz}
\author[ $\dag$]{Alessandro Calefati}
\affil[*]{ Department of Theoretical and Applied Science, University of Insubria, Varese, Italy }
\affil[$\dag$]{ Department of Theoretical and Applied Science, University of Insubria, Varese, Italy }


%


\maketitle

\begin{abstract}
%
With massive explosion of social media such as Twitter and Instagram, people daily share billions of multimedia posts, containing images and text.
Typically, text in these posts is short, informal and noisy, leading to ambiguities which can be resolved using images. 
In this paper we explore text-centric Named Entity Recognition task on these multimedia posts. We propose an end to end model which learns a joint representation of  a text and an image. 
Our model extends multi-dimensional self attention technique, where now image  help to enhance relationship between words. Experiments show that our model is capable of capturing both textual and visual contexts with greater accuracy, achieving state-of-the-art results on Twitter multimodal Named Entity Recognition dataset.


\end{abstract}


%
\IEEEpeerreviewmaketitle

\section{Introduction}
\begin{figure}[t]
  \centering 
  \begin{footnotesize}
  \begin{tabular}{p{2.7cm}p{5.1cm}}
  \includegraphics[height=0.19\textwidth]{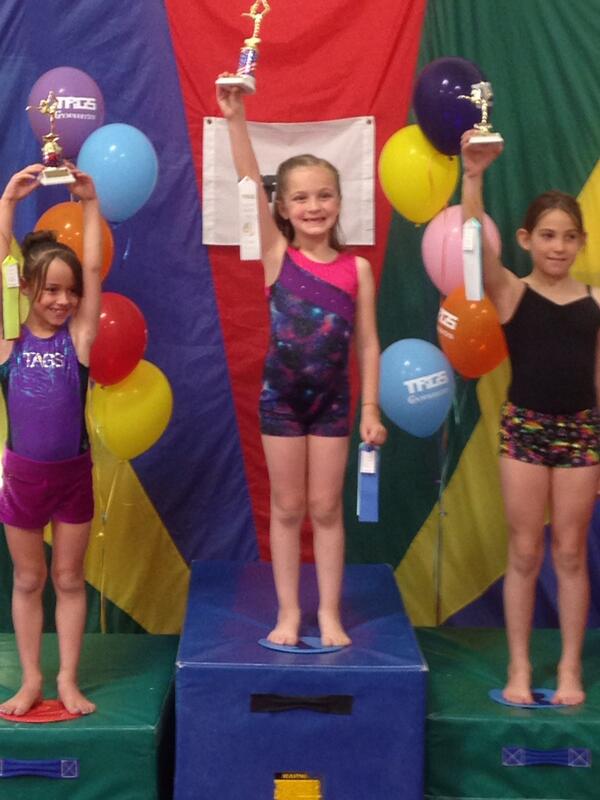} & 
  \includegraphics[height=0.19\textwidth]{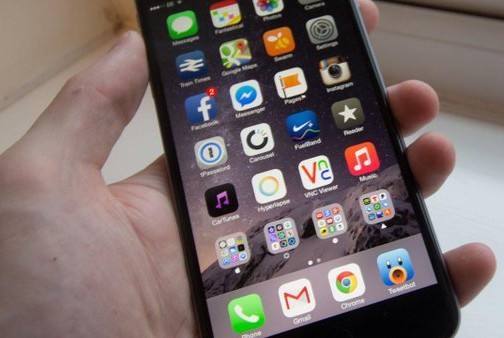}\\
My daughter got 1 place in [Apple valley \textbf{LOC}] Tags gymnastics & 
[Apple \textbf{ORG}] 's latest [iOS \textbf{OTHERS}] update is bad for advertisers \\

  \end{tabular}
  \end{footnotesize}
  \caption{Two NER multimodal examples show how some entities in the text can be correctly tagged in combination with visual information. 
  Looking only at the text, the word \textit{Apple} is ambiguous in the text description on the left, because it can be interpreted as \textit{Location} (LOC) or as \textit{Organization} (ORG).}
  \label{fig:the-problem}
\end{figure}

Recent years have seen a surge in multimodal data containing various media types. Typically, users combine image, text, audio or video data to express views on a social media platform. 
The combination of these media types has been extensively studied to solve various tasks including classification~\cite{arevalo2017gated,kiela2018efficient,gallo2017multimodal}, cross-modal retrieval~\cite{nawaz2018revisiting,wang2016learning} semantic relatedness~\cite{kiela2014learning,leong2011going} and Visual Question Answering (VQA)~\cite{fukui2016multimodal,anderson2018bottom}.  
Recently, works in~\cite{zhang2018adaptive, moon1078} and~\cite{lu2018visual} combined text and image in a multimodal approach for text Named Entity Recognition (NER) problem~\cite{baldwin2015shared}.
Typically, the text component of a NER multmodal problem is challenging due to  informal language, slang and typos etc.~\cite{baldwin2015shared}.
These attributes make the task more challenging, compared to traditional NER.
Moreover there are some ambiguous cases that can only be resolved  with visual context, shown in Fig.~\ref{fig:the-problem}.
If we consider only the text in the first example ``My daughter got $1$ place in Apple valley Tags gymnastics'', \textit{Apple} is recognized as the name of an \textit{Organization}, but, in this tweet, \textit{Apple} should be labeled as \textit{Location}.
Similarly, the text in the second example ``Apple's latest iOS update is bad for advertisers'', \textit{Apple} is wrongly recognized as the name of an \textit{Organization}.
In both cases, the disambiguation of the text is a non-trivial task, without considering the visual context.

In this work, we propose a novel neural network architecture which leverages on visual context to recognize named entities.
We combine character and word embeddings to handle characteristics of NER textual component.
In addition, self attention mechanism is extended to capture relationships between two words and image regions, unlike previous works~\cite{zhang2018adaptive,lu2018visual} which used only single words to capture visual attention.  
Finally, we introduce a gated multimodal fusion module to select information dynamically from textual and visual features.
Intuitively, our model captures two forms of interactions: intra-modal and cross-modal interactions.\footnote{Intra-modal interactions deal within same modality, whereas cross-modal captures interaction between modalities.}
We achieved state-of-the-art results on NER multimodal dataset~\cite{zhang2018adaptive}.
In addition, we performed extensive experiments to show the effectiveness of the proposed model. 
Our main contributions are:

\begin{itemize}
    \item introduction of an end-to-end model based on attention only that jointly learns intra and cross-modal dependencies, enhancing relationship of two words;
    \item state-of-the-art results on NER multimodal dataset~\cite{zhang2018adaptive}.
\end{itemize}

\section{Related work}
In this section, we summarize relevant background on previous works on attention techniques and multimodal NER.\\
\textbf{Attention Techniques.} 
Attention techniques allow models to focus on parts of visual or textual inputs of a task.
Visual attention models selectively pay attention to small regions of an image to extract features.
On the other hand, textual attention techniques find semantic or syntactic  alignments in handling long-term dependencies.
Attention techniques have been extensively employed to vision and text related tasks, such as Image Captioning~\cite{xu2015show}, VQA~\cite{anderson2018bottom}, Cross-Modal Retrieval~\cite{nam2017dual} NER ~\cite{zhang2018adaptive, moon1078, lu2018visual}.  
\\
\textbf{Multimodal Named Entity Recognition.} NER task for short and noisy texts has been extensively studied in the literature~\cite{lample2016neural, aguilar2017multi}. 
Recent years have seen an interest in capturing visual context from social media posts to improve this task~\cite{zhang2018adaptive, moon1078, lu2018visual}. 
These works used bi-directional Long Short Term Memory (LSTM) networks to extract features from a sequence of words.
The work in~\cite{zhang2018adaptive} captures interactions between words and an image in a bi-directional way.
However, it represents this interaction in a uni-directional manner.
In our work, we extended multi-dimensional attention to jointly learn intra and cross-modal dependencies.

\section{Proposed Model}
In this paper we propose a novel architecture, inspired from Disan~\cite{shen2017disan}, to learn a joint representation of text and image for multimodal NER.
Our model improves the intra-text attention to learn enhanced representations exploiting the relevancy with images. 
Instead of learning text representations separately from textual context and then leveraging on image information~\cite{zhang2018adaptive} and~\cite{lu2018visual}, our model jointly learns shared semantics between intra-text representation and visual features.
In next subsections, we first explain each module of our network and then, the proposed end to end model, shown in~Fig.~\ref{fig:proposed-model}.

\subsection{Attention}
\begin{figure}
  \centering 
  \includegraphics[width=0.95\columnwidth]{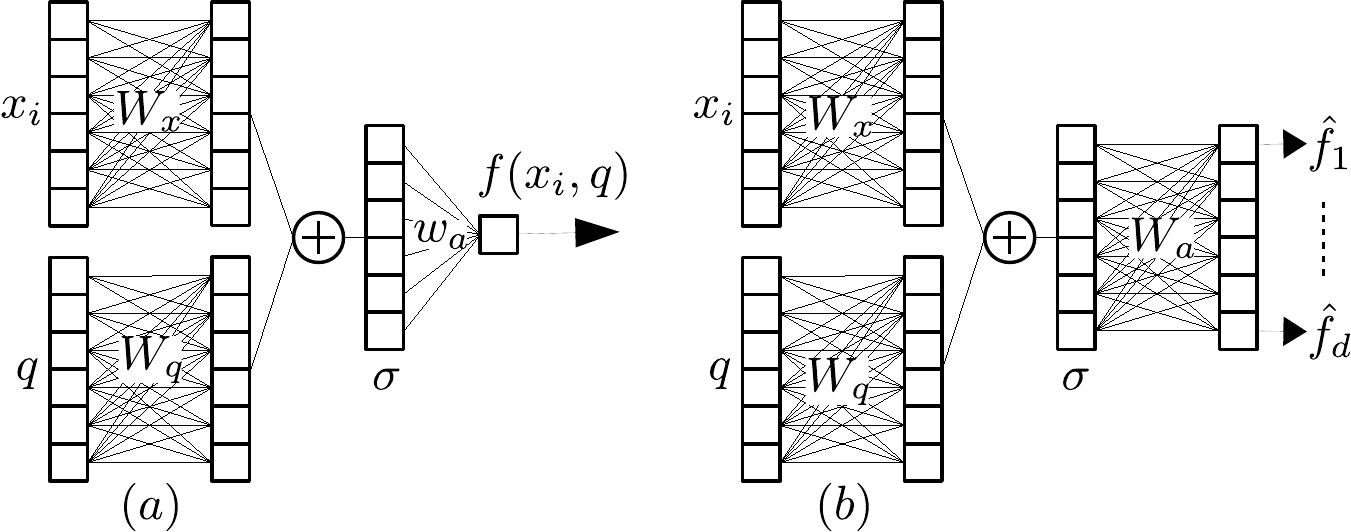} 
  \caption{
  (a) A neural representation of the additive attention described in Eq.~(\ref{eq:additive_attention}), for two word embeddings $q$ and $x_i$. (b) The additive multi-dimensional attention described in Eq.~(\ref{eq:alignment_score_multidim}).
  }
  \label{fig:additive_attention}
\end{figure}

The purpose of the attention module is to compute an \textit{alignment score} between elements coming from different sources. 
In Natural Language Processing, given a sequence of word embeddings ${\textbf{x}} = [{x}_{1},{x}_{2},...,{x}_{n}]$ and the embedding of a query $q$, with $x_i, q \in \mathbb{R}^{d_e}$, the alignment score between $x_i$ and $q$ can be calculated using the common \textit{additive attention}:
\begin{equation}\label{eq:additive_attention}
  f(x_i,q) = w_a \sigma(x_i W_x + q W_q)
\end{equation}
where $\sigma$ is an activation function, $w_a$ is a vector of weights and $W_q$, $W_x$ are weights matrices. A graphical representation of the additive attention is available in Fig.~\ref{fig:additive_attention}(a).
Furthermore, we use a special case of attention called \textit{self attention} in which both elements $q$ and $x_i$ come from the same source.

$f(x_i,q)$ is a scalar score, which determines how important $x_i$ is to a query $q$. 
Alignment score between $q$ and all tokens becomes:
\begin{equation}\label{eq:alignment_score}
  a = [f(x_1,q),\dots, f(x_n,q)]
\end{equation}

Then, a probability distribution $p(z\mid \textbf{x},q)$ is calculated over $a$ by applying softmax. 
This gives a measure on how much the token $x_i \in \textbf{x}$ is important to a query $q$.
\begin{equation}\label{eq:alignment_score_prob}
  p(z\mid \textbf{x},q) = softmax(a)
\end{equation}

The final output we obtain from attention vector $a$ is the weighted sum of all tokens in $\textbf{x}$ 
\begin{equation}
  C = \sum_{i=1}^{n} p(z = i\mid \textbf{x},q) x_i
\end{equation}
that is the \textit{context vector} for a query $q$.

\subsection{Multi-Dimensional Attention}

Multi-dimensional attention~\cite{shen2017disan} proposed for self attention, is an additional attention technique which computes the feature-wise score vector $\hat{f}(x_{i},q)$ described in Eq.~(\ref{eq:alignment_score_multidim}) instead of computing the scalar score shown in Eq.~(\ref{eq:additive_attention}). 

\begin{equation}\label{eq:alignment_score_multidim}
\hat{f}(x_{i},q) = W_a \sigma(x_i W_x  +  q W_q )
\end{equation} 

where $\hat{f}(x_i,q) \in \mathbb{R}^{d_e}$ is a vector with the same length  as $x_i$, and $W_a$, $W_q$, $W_x \in \mathbb{R}^{d_e \times d_e}$ are the weight matrices.

Softmax is applied to the output function $\hat{f}$ to compute categorical distribution $p(z \mid \textbf{x},q)$ over all tokens.

To find the importance of each feature $k$ in a word embedding $x_i$, $\hat{f}(x_i,q)$ becomes $[\hat{f}(x_i,q)]_{k}$ and the categorical distribution is calculated as:

\begin{equation}\label{eq:categorical_distribution}
 P_{ki} = p(z_{k} = i \mid \textbf{x},q) = softmax([\hat{f}(x_i,q)]_k)
\end{equation}

Thus the final context becomes:
\begin{equation}\label{eq:Final_output}
 C = \Big[\sum_{i=1}^{n} P_{ki} x_{ki} \Big]_{k=1}^{d_e}
\end{equation}

The \textit{multi-dimensional attention} defined in Eq.~(\ref{eq:alignment_score_multidim}) is known as "token2token" self attention.
It explores dependency between elements of the same source, i.e. query $q$ and word $x_i$ from a single source $\textbf{x}$.

\subsection{Image Feature extraction}
In order to obtain features $F$ from an image $I$ we use a pretrained VGG-19 model. 
We extracted features of different image regions from the last pooling layer which has a shape of $7\times7\times512$.
To simplify the calculations, we resized it to $49\times512$.
We have $N=49$ regions with $d_i=512$ as dimension for each feature vector $F_j$ with $j \in [1,\dots,N]$.
Regional features of a given image are represented by the matrix $F = VGG(I)$.

\subsection{Character-based representation}
Text extracted from social media is usually informal.
In addition, it contain many out of vocabulary words.
Character level features can play a crucial role in handling such text.
We use a 2D Convolutional Neural Network to learn character embeddings. 
A word $w$ is transformed into a sequence of characters $c = [c_1,c_2,...,c_n]$ where $n$ is the word length. 
A convolutional operation of filter size $1\times k$ is applied to the matrix $W$, where $W \in \mathbb{R}^{d_e\times n}$ and $d_e$ is the character embedding size. 
Then, we compute column-wise maximum operation to get the embedding for word $w$.

\subsection{Attention Guided Visual Attention}
\begin{figure}
  \centering
  \includegraphics[width=0.85\columnwidth]{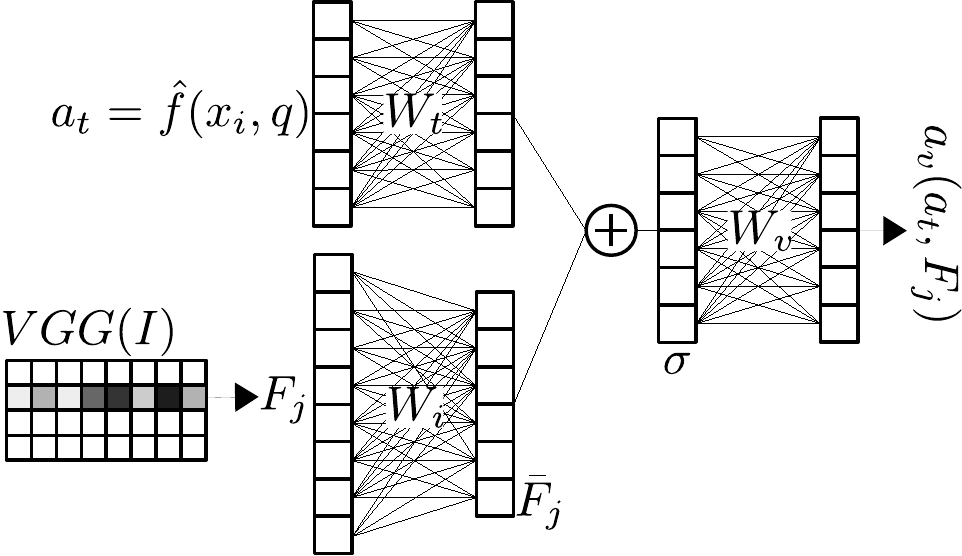}
  \caption{Attention Guided Visual Attention neural representation.
  The output of this model is a single vector that will be concatenated to the others to obtain the final matrix.
  }
  \label{fig:attention-guided-visual-attention}
\end{figure}

Our proposed model uses alignment score between two textual tokens to compute cross-modal attention, show in Fig.~\ref{fig:attention-guided-visual-attention}. 
Given a word and query $q$, their alignment score is calculated using Eq.~(\ref{eq:alignment_score_multidim}).
\begin{equation} \label{eq:alignment_score_text}
 a_{t} = \hat{f}(x_i, q)
\end{equation} 
where $a_{t}$ is a feature-wise score vector with same length of $x_i$. 
We calculate attention between $a_{t}$ and image feature matrix $F$.
\begin{equation}\label{eq:attention}
 a_v(a_t, F_j)= W_v \sigma (a_t W_t  +  F_j W_i   ) 
\end{equation}
where $a_v(a_t, F_j)$ is a vector representing a single row of the visual attention scores matrix between $a_t$ and $F$, $a_t \in \mathbb{R}^{d_e}$, $F_j \in \mathbb{R}^{d_i \times N}$, $N$ is number of image regions, 
$W_i \in \mathbb{R}^{d_i \times d_e}$, $W_t$, $W_v \in \mathbb{R}^{d_{e} \times d_{e}}$ are the weight matrices.
To obtain the final visual attention matrix, we compute $a_v$, such that $a_v$ $\in \mathbb{R}^{d_e \times N}$

Then we normalize the scores $a_v$ by applying Eq.~(\ref{eq:categorical_distribution}) to get a probability distribution (columns-wise) over all regions of image.
\begin{equation}\label{eq:softmax}
  P(a_v) = softmax(a_v) 
\end{equation}

\begin{figure}
  \centering
  \includegraphics[width=0.95\columnwidth]{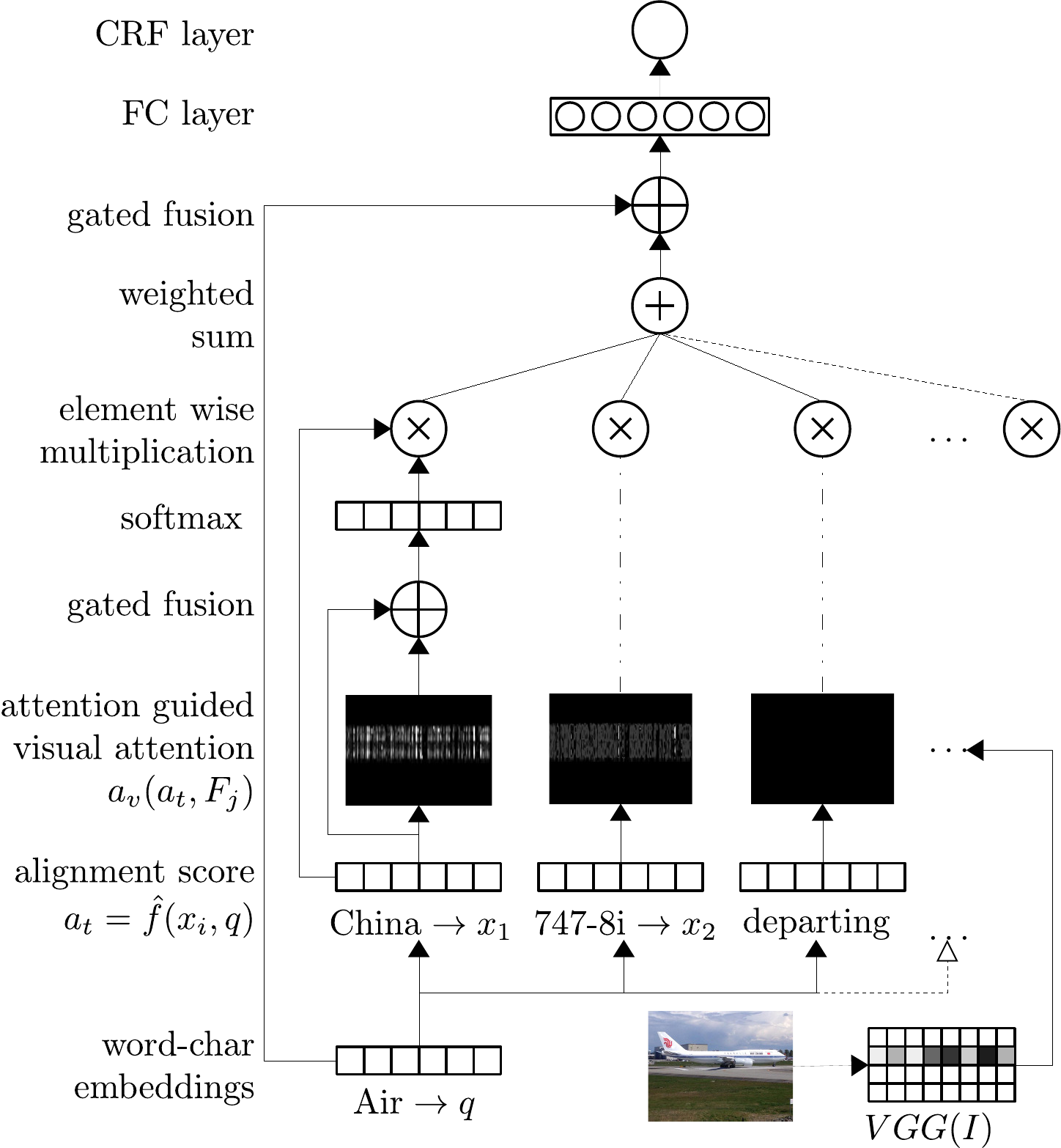}
  \caption{
  End-to-end Multi Dimensional Attention based model improving intra-modal attention using visual context for multimodal NER.
  }
  \label{fig:proposed-model}
\end{figure}

The final output is a element-wise product between $p(a_v)$ and $F$.

\begin{equation}\label{eq:attention_final_output}
  C_v = \sum_{i=1}^{n} P_i(a_v)  \odot F_i
\end{equation}
where  $C_v \in \mathbb{R}^{d_e}$, is a vector containing context vector for $a_t$.

\subsection{Gated Fusion}
In order to combine alignment score $a_t$ of a word $w$ and a query $q$ with its visual attention vector $C_v$ we use a gate function to dynamically combine alignment score and visual attention vectors.

\begin{equation}\label{eq:gate}
  G = \sigma (W^{(1)} a_t +  W^{(2)} C_v  + b^{(G)} ) 
\end{equation}
\begin{equation}\label{eq:gate2}
  O = G \odot C_v + (1 - G) \odot  a_t
\end{equation}
Throughout our network, we used gated fusion to merge different modalities.

\section{End to End Model}
Our end to end network jointly learns intra and cross-modal dependencies. 
The architecture is shown in Fig.~\ref{fig:proposed-model}.

Given a sequence of words $\textbf{x} = [x_1, x_2,\dots ,x_n]$, and image features $F$, we first compute the alignment score between $x_i$ and a query $q$,  where $i$ ranges from $1$ to the length $n$ of a sentence.
We use Eq.~(\ref{eq:alignment_score_text}) to compute the alignment score $a_t$.
%
In order to aid intra-text representation $a_t$, we compute alignment score $alignment_v$ between $a_t$ and $F$ using compatibility function given in Eq.~(\ref{eq:attention}).

\begin{equation}
 alignment_v = a_v(a_t, F)
\end{equation}

We then apply the method described in Eqs.~(\ref{eq:softmax})-(\ref{eq:attention_final_output}) on $alignment_v$ to get a weighted sum $C_v$. 
Now this $C_v$ is a visual context vector, which will contribute to $a_t$. 
This step makes intra-modal relation $a_t$ between $x_i$ and $q$ stronger because it also includes image context. 
Relation $a_t$ between two words will remain the same if textual context is taken into account. 
But if we included visual context, this relation would become dynamic and more expressive. 
Thus helping intra-modal attention to learn improved relationships between words.

In order to make visual context vector $C_v$ helpful for $a_t$, we combine them using gated fusion.
This will dynamically select which features to select from various modalities. 
We applied Eqs.~(\ref{eq:gate})-(\ref{eq:gate2}) to get a fused representation $F_{rep}$ between $a_t$ and $C_v$.

Now we have fused representation $F_{rep}$ representing $a_t$ between query $q$ and $x_i$. Note that, this is calculated for all tokens of a sequence $\textbf{x}$.
We apply Eq.~(\ref{eq:categorical_distribution}) to get a categorical distribution $P$ over all $n$ tokens of $\textbf{x}$. 
Then, element-wise product is computed between $P$ and each token of a sequence $\textbf{x}$ to get context vector $C$ for query $q$.
\begin{equation}
 C = \sum_{n=1}^{n} P_i  \odot x_i
\end{equation}
where $C$ $\in \mathbb{R} ^ {d_e}$, $n$ is total tokens of a sentence and $d_e$ is vector dimension of each token. Now $C$ is a context vector jointly learned from text and visual features. To handle attributes of text component of NER, we also used word representation $x_i$ with $C$, exploiting gated fusion Eqs.~(\ref{eq:gate})-(\ref{eq:gate2}) to obtain the final output $O$. 
Furthermore, $O$ is passed through a fully connected layer.
\begin{equation}\label{eq:FC}
 O_{fc} = Relu(W^{(1)} O + b_o )
\end{equation}
where $W^{(1)}$ is the learnable parameter and $b_o$ is the bias vector.
For tag prediction, final output $O_{fc}$ is passed to Conditional Random Field (CRF) layer~\cite{MaH16}.

\subsection{Conditional Random Field}
CRF~\cite{Lafferty} are useful in tasks where output labels have a strong dependency (e.g. I-PER cannot follow B-LOC). Predicting such outputs independently is challenging without correlation between labels and their neighborhood. 
Given ${\textbf{x}}  = [{x}_{1},{x}_{2},...,{x}_{n}]$ as a text sequence and ${\textbf{y}}  = [{y}_{1},{y}_{2},...,{y}_{n}]$ as a sequence of labels for ${\textbf{x}}$, possible labels sequences can be calculated using following equation.
\begin{equation}
p(\textbf{y}\mid \textbf{x}) = \frac{\displaystyle \prod_{i=1}^{n}\Omega_i(y_{i-1},y_i,\textbf{x})} {\displaystyle \sum_{y^\prime \in Y}{×}\displaystyle \prod_{i=1}^{n}\Omega_i(y^\prime_{i-1},y_i^\prime,\textbf{x})}
\end{equation}
Where $\Omega_i(y_{i-1},y_i,\textbf{x})$ and $\Omega_i(y^\prime_{i-1},y_i^\prime,\textbf{x})$ are potential functions. We use maximum conditional likelihood to learn best parameters that maximize the log-likelihood.

\begin{equation}
 L(p(\textbf{y}\mid \textbf{x})) = \sum_{i}logp(\textbf{y}\mid \textbf{x}))
\end{equation}

\section{Datasets}
\begin{figure}
\centering
\begin{tabular}{p{3.5cm}p{4.5cm}}
\includegraphics[height=0.14\textwidth]{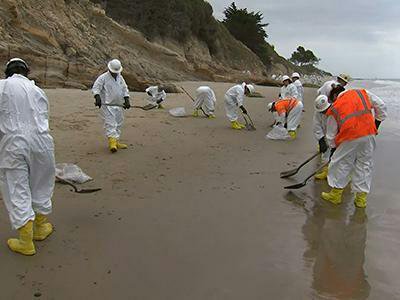} & \includegraphics[height=0.14\textwidth]{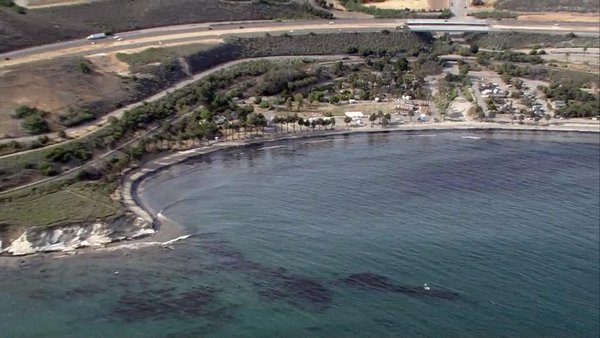} \\
(a) Finding [California \textbf{OTHER}] [oil \textbf{OTHER}] [spill \textbf{OTHER}]`s cause could take a month &
(b) Oil pipeline break dumps crude oil on California \textbf{[LOC]} beach
\end{tabular}
\caption{Some examples showing how images can help to resolve ambiguities. 
The word "California" in (a) and (b) is difficult to understand without looking at the images as it has different tags.}
\label{fig:dataset}
\end{figure}

We used multimodal NER dataset~\cite{zhang2018adaptive}.
It contains $4$ types of entities \{Person, Location, Organization and Misc.\} collected from $8257$ tweets,
containing $4000/1000/3257$ samples for training/val/test sets respectively.
Table~\ref{Table:dataset} shows number of samples per entity.
Fig.~\ref{fig:the-problem} and ~\ref{fig:dataset} show some ambiguous examples.


\begin{table}[htbp]
\caption{Details of Dataset}
\begin{center}
\begin{tabular}{c|c|c|c}
 \hline
& \textbf{Train} &\textbf{Val} & \textbf{Test} \\
\hline\hline
 Person      & 2217  & 552  & 1816  \\
\hline
Location     & 2091  & 522  & 1697    \\
\hline
Organization & 928  & 247  & 839       \\
\hline
Misc         & 940  & 225  & 726         \\
\hline
\end{tabular}
\label{Table:dataset}
\end{center}
\end{table}

\section{Experiments}
We performed various experiments to evaluate the effectiveness of the proposed model on NER multimodal dataset~\cite{zhang2018adaptive}.
Standard Precision, Recall and F1 scores are used as evaluation metrics.

\subsection{Baselines}
\subsubsection{Disan}
Our model is inspired from Disan, thus we consider this approach as a baseline for the text only evaluation. 
We used multi-dimensional self attention method to extract context-aware representation of texts. 
For fair comparison, we keep the same architecture of Fig.~\ref{fig:proposed-model}, but we exclude the ``Attention Guided Visual Attention'' module.
This to analyze differences between the proposed model and multi-dimensional self attention approaches.

\subsubsection{State-of-the-art models}
We compare our model with previous state-of-the-art methods leveraging on LSTM networks to capture textual dependencies and visual attention to exploit cross-modal interactions.

\subsection{Word embeddings}
We used 300D fasttext crawl embeddings. 
It contains $2$ million word vectors trained with subword information on Common Crawl ($600$B tokens). 
However, we do not apply fine-tuning on these embeddings during the training stage.

\subsection{Character Embeddings}
$50$D character embeddings are trained from scratch using a single layer 2D CNN with a kernel size of $1\times3$.

\subsection{Optimization}
We set Adam optimizer with different learning rate initialization: 
$0.001$, $0.01$ and $0.005$. 
We achieved the best score using the learning rate equal to $0.001$.
Batch size iand dropout keep probability are set $20$ and $0.5$ respectively.

\begin{figure}[h]
  \centering
\begin{tabular}{ccl}
 \includegraphics[width=0.21\textwidth]{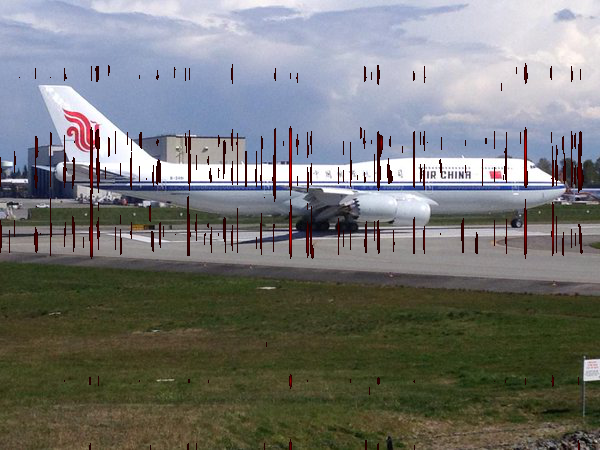} &  
 \includegraphics[width=0.21\textwidth]{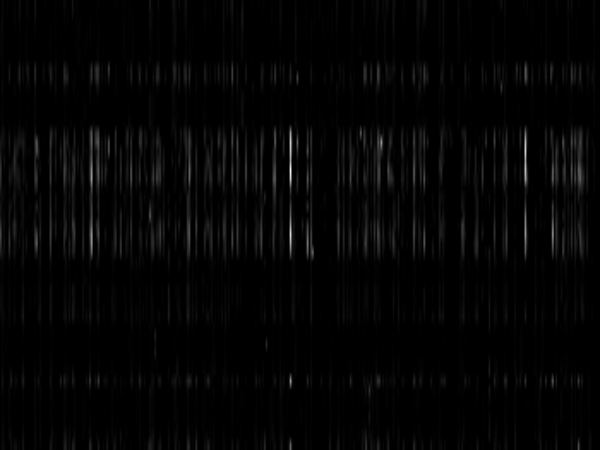} & \hspace{-0.3cm}(a)\\
 \includegraphics[width=0.21\textwidth]{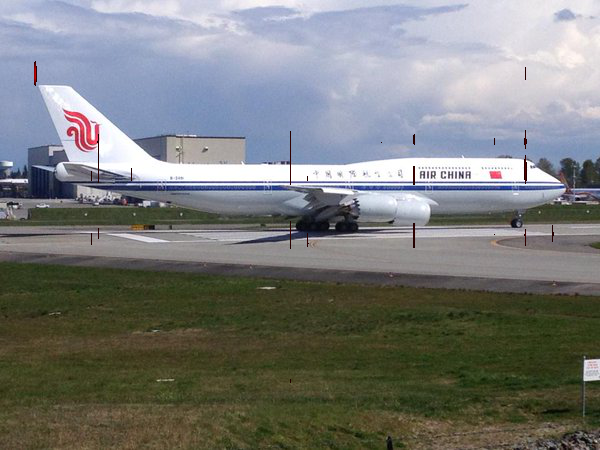} &  
 \includegraphics[width=0.21\textwidth]{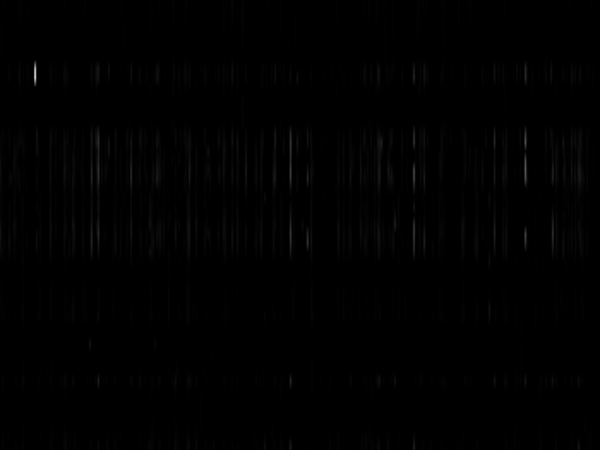} & \hspace{-0.3cm}(b)\\
 \includegraphics[width=0.21\textwidth]{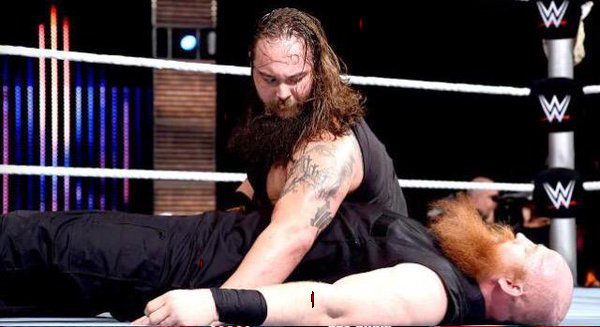} &
  \includegraphics[width=0.21\textwidth]{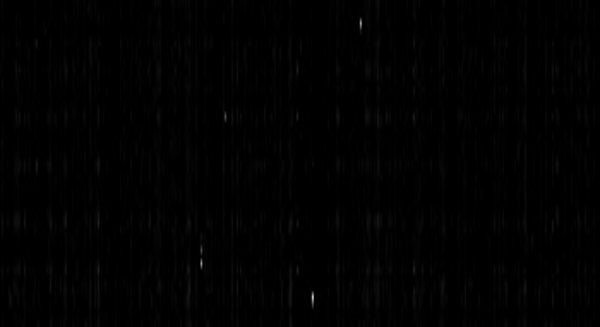} & \hspace{-0.3cm}(c)
\end{tabular}
  \caption{
  Examples of ``attention guided visual attention''. In (a) a graphical representation of the visual attention matrix $a_v(a_t, F_j) \forall j$, for the the alignment score $a_t=\hat{f}(x_i,q)=\hat{f}(\text{Air}, \text{China})$ and in (b) a graphical representation of the visual attention matrix for  the alignment score $a_t=\hat{f}(x_i,q)=\hat{f}(\text{Air}, \text{Field})$. In (c) the visual attention matrix obtained from a different image.
  }
  \label{fig:attention_guided_viasual}
\end{figure}

\section{Results}
Table~\ref{Table:results} shows comparison of our model with baseline and previous state-of-the-art methods. 
We achieved best $F1$ score, outperforming previous approaches.

\subsection{Impact of word embeddings}
Word embeddings with different sizes and trained on different corpora have a strong impact on performance. 
We evaluate the model on two word embeddings to analyze the effect.

\begin{itemize}
\item Twitter Embeddings: we used embeddings trained on $30$ million tweets~\cite{zhang2018adaptive}. 
To compare our results with this work, we used same embeddings to evaluate the improvement that comes from our proposed architecture.
\item Crawl Embeddings: we used $300$D Crawl embeddings~\cite{grave2018learning} trained on $600$B tokens to fully exploit the capability of our model, 
 
\end{itemize}
Table~\ref{Table:results} shows results with different word embeddings. 
Our model achieves better results using Twitter Embedding, but there is a clear performance boost with Crawl embeddings.
It is interesting to note that entities ``PER'' and ``LOC'' have a minor improvement whereas there is an enhancement of 5-6\% in ``ORG'' and ``MISC''. 
With Crawl embeddings, our model performs best to predict the ``ORG'' category, with a 3\% difference from  baseline (text only using the Disan model). 
Our model achieves best F1 scores using both embeddings.
This clearly shows that our model performs better regardless to the word embedding.

\subsection{Impact of cross-modal attention}
Table~\ref{Table:results} shows that ``Attention guided visual attention'' boosts Disan (intra-modal attention) performance. 
We present a qualitatively example in Fig~\ref{fig:attention_guided_viasual} that shows the focus of visual attention.
Given an annotated sentence:\\

\textbf{S} = [Air B-OTHER] [China I-OTHER] [747-8i I-OTHER] [departing O]... [Field O] \\

an alignment score (inter-modal attention) $a_t = \hat{f} (x_i,q)$ between two tokens ``Air'' and ``China'' is aided by visual context, see Fig.~\ref{fig:attention_guided_viasual} (a).
Our model can successfully focus on related image regions, strengthening the relation between two words. 
Whereas alignment score  $a_t = \hat{f} (x_i,q)$ between ``Air'' and ``Field'' and visual context (Fig.~\ref{fig:attention_guided_viasual} (b)) has less relation because word ``Air'' has higher dependency on ``China'' then on ``Field'' when labeling ``Air'' as ``B-OTHER''. 
Similarly in order to verify that relation between two tokens varies according to image, we changed the image and kept same sentence. 
Relation between fake image and alignment score $a_t =\hat{f} (x_i,q)$ between two tokens ``Air'' and ``China`` can be seen in Fig.~\ref{fig:attention_guided_viasual}(c) with no prominent relation, proving that our model learns better relationships between words given an image.

\begin{figure}
\centering
\begin{tabular}{p{3.3cm}p{4.9cm}}
\includegraphics[height=0.18\textwidth]{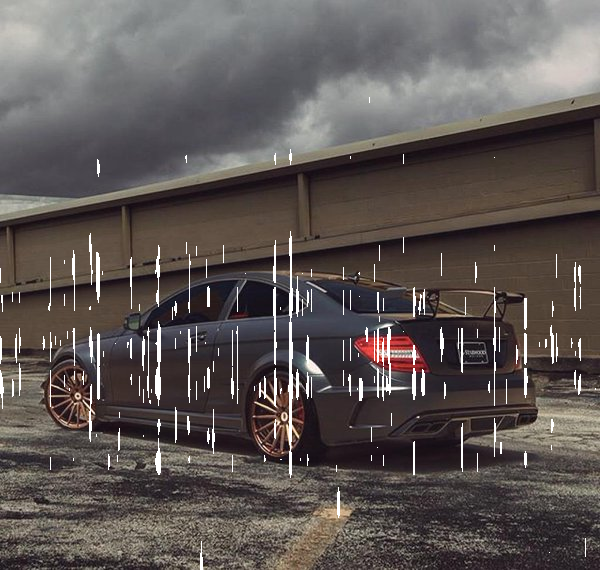} &
\includegraphics[height=0.18\textwidth]{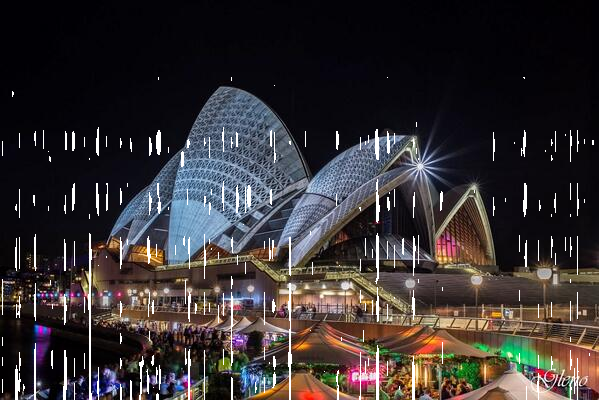} \\
(a)~$[$Mercedes~\textbf{OTHER}$]$ $[$Benz~\textbf{OTHER}$]$ & 
(b)~photo~of~$[$Sydney~\textbf{LOC}$]$ $[$Opera~\textbf{OTHER}$]$~$[$House~\textbf{OTHER}$]$ 
\end{tabular}
\caption{Two example of correct visual attention. Our model successfully highlights related image regions required to predict correct tag.}
  \label{fig:correct_Examples}
\end{figure}

Similarly, Fig.~\ref{fig:correct_Examples} shows some examples in which our model pay attention to related image regions. 
It clearly shows that our model focuses on certain parts of image which are beneficial for named entity task. 
Fig.~\ref{fig:correct_Examples}(a) shows that attention is payed only to car to predict correct tag for ``Mercedes'' and ``Benz''. Similarly Fig.~\ref{fig:correct_Examples}(b) shows that ``Opera House'' is a building and not a location, and our models correctly identifies it by paying attention to correct image regions.

\begin{table*}[t]
\caption{Comparison of our approach with baselines and previous state-of-the-art methods.}
\centering
\resizebox{0.85\textwidth}{!}{%
\begin{tabular}{c|c|c|c|c|c|c|c}
\hline
& \multicolumn{1}{c|}{\begin{tabular}[c]{@{}c@{}}PER\\ F1\end{tabular}} & \multicolumn{1}{c|}{\begin{tabular}[c]{@{}c@{}}LOC\\ F1\end{tabular}} & \multicolumn{1}{c}{\begin{tabular}[c]{@{}c@{}}ORG\\ F1\end{tabular}} & \multicolumn{1}{c|}{\begin{tabular}[c]{@{}c@{}}MISC\\ F1\end{tabular}} & \multicolumn{3}{c}{\begin{tabular}[c]{@{}c@{}}Overall\\ Prec. Recall F1\end{tabular}} \\ \hline \hline
T-NER ([11) & 83.64   & 76.18   & 50.26  & 34.56   & 69.54  & 68.65 & 69.09  \\ \hline
Adaptive Co-Attention Network \cite{zhang2018adaptive}   & 81.98   & \textbf{78.95}  & 53.07 & 34.02 & 72.75  & 68.74  & 70.69 \\ \hline
Disan~\cite{shen2017disan} (Twitter Embedding)  & 82.07  & 76.87      & 55.34    & 32.29  & 71.00    & 70.53  & 70.77  \\ \hline
Our Model (Twitter Embedding)     & 82.83  & 78.22  &  55.88   & 33.00  & 72.81     & 70.33    & 71.55    \\ \hline
Disan \cite{shen2017disan} (Crawl Embedding) & 83.03    & 77.96   & 56.66   & \textbf{39.54}   & 71.65   & 71.89     & 71.77     \\ \hline
Our Model  (Crawl Embedding)    & \textbf{83.98}     & 78.65    & \textbf{59.27}     & \textbf{39.54}      & \textbf{73.50}  &  \textbf{72.33 } & \textbf{72.91}  \\ \hline
\end{tabular}%
}
\label{Table:results}
\end{table*}

\subsection{Error Analysis}
\begin{figure}[h]
 \centering
 \begin{tabular}{p{4.0cm}p{4.0cm}}
 \includegraphics[height=0.13\textwidth]{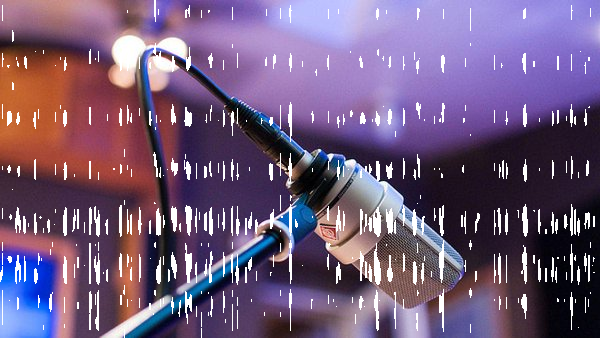} &
 \includegraphics[height=0.13\textwidth]{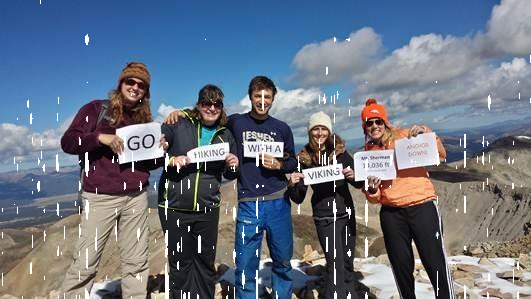}\\
 (a) $[$Reddit~\textbf{ORG}$]$ needs to stop pretending &
 (b) teachers on top of $[$Mount~\textbf{PER}$]$ $[$Sherman~\textbf{PER}$]$
 \end{tabular}
  \caption{ Two examples of wrong visual attention:
  (a) shows an unrelated image and a wrong prediction, while 
 (b) shows a related image with wrong attention and prediction.
 }
 
  \label{fig:wrong_Examples}
\end{figure}

In Fig.~\ref{fig:wrong_Examples}, we show some examples where our approach fails because of the following reasons:

\begin{itemize}
 \item Unrelated image : Text information do not match with an image, as we can see in Fig.~\ref{fig:wrong_Examples}(a), "Reddit" belongs to "Other" but unrelated image caused wrong attention and it results to wrong prediction "ORG".
 
 \item Wrong attention: Fig.~\ref{fig:wrong_Examples}(b) shows an example where text and image are aligned correctly, but wrong attention results in wrong tag prediction. 
 Words "Mount" and "Sherman" were tagged as "Person" as most of attention is on persons, whereas expected tag was "LOC".
\end{itemize}

\section{Conclusion}
We introduced a novel model for multimodal NER.
It extends multi-dimensional self attention approaches enhancing intra-text relationships using visual features.
Qualitative examples show that our model successfully captures correct relations between words and images removing ambiguities caused by the text.
Our model is flexible and can be further extended to other multimodal tasks.
Experiments show that our model achieved state-of-the-art results on multimodal NER dataset.

\bibliographystyle{IEEEtran}
\bibliography{IEEEabrv,icdar}

\end{document}